# Deep clustering of longitudinal data

Louis Falissard*

Oxford University, Nuffield Department of Surgical Sciences, Computational Neuroscience lab

Guy Fagherazzi

Inserm U1018, Gustave Roussy Institute, CESP, Villejuif, France

University Paris-Saclay, University Paris-Sud, Villejuif, France

Newton Howard

Oxford University, Nuffield Department of Surgical Sciences, Computational Neuroscience lab

Bruno Falissard

CESP, INSERM U1018, Université Paris-Sud, Université Paris-Saclay, UVSQ, Paris, France

Abstract: Deep neural networks are a family of computational models that have led to a dramatical improvement of the state of the art in several domains such as image, voice or text analysis. These methods provide a framework to model complex, non-linear interactions in large datasets, and are naturally suited to the analysis of hierarchical data such as, for instance, longitudinal data with the use of recurrent neural networks.

In the other hand, cohort studies have become a tool of importance in the research field of epidemiology. In such studies, variables are measured repeatedly over time, to allow the practitioner to study their temporal evolution as trajectories, and, as such, as longitudinal data. This paper investigates the application of the advanced modelling techniques provided by the deep learning framework in the analysis of the longitudinal data provided by cohort studies. A method for visualizing and clustering longitudinal dataset is proposed, and compared to other widely used approaches to the problem on both real and simulated datasets. The proposed method is shown to be coherent with the preexisting procedures on simple tasks, and to outperform them on more complex tasks such as the partitioning of longitudinal datasets into non-spherical clusters.

# Introduction

The question of longitudinal data analysis in cohort studies is a topic of importance in epidemiology. In this particular type of study, variables are observed repeatedly over long periods of time. As a consequence, a proper statistical analysis of the gathered data needs to account for the variable's evolution through time just as well as its value.

Sequence clustering procedures have shown to perform particularly well in that task. The main idea behind cluster analysis is to group individuals together according to some predetermined similarity criterion, such that the resulting groups of individuals all display similar properties, which can then be used by the analyst to identify and extract useful insight from the data.

Even though the notion of similarity between individuals was traditionally designed to exploit geometrical properties of multivariate, non-temporal datasets, it has been successfully expanded to longitudinal datasets through a range of measurements, such as, for instance, temporal Euclidian distance.

Several clustering procedures have been developed for the analysis of longitudinal data. The kml package introduced in the open source statistical programming language R, for instance, relies on a K-mean algorithm coupled with a longitudinal distance measurement, which is traditionally defined as the aforementioned temporal Euclidian distance.

Another method, proc traj, was introduced in the software SAS. This approach differs significantly from the kml K-mean, and offers a likelihood approach based on gaussian mixture of parametric models assumed to for the phenomenon to be studied. Both approaches present advantages and drawbacks, and reasons to pick amongst the different available methods are not well defined.

Even though sequence clustering is approached as a statistical analysis problem in epidemiology, most of its underlying theory shares deep roots with the growing field of machine learning. More specifically, unsupervised machine learning, which consists of learning specific structures in a dataset, sees clustering algorithms as one of its biggest field. Therefore, it is natural to investigate the question of longitudinal clustering within a machine learning approach.

In the past few years, the field of machine learning has been subject to a significant expansion, leading to the development of widely used applications in many field such as, for instance, image analysis, voice analysis, or natural language processing. Those recent progress are mainly due to the recent successes of a machine learning subfield, deep learning, in modelling complex, structured data, through the construction of an increasingly complex, hierarchical representation of a dataset. More specifically, the artificial neural network approach proposed in deep learning provides a framework naturally able to model longitudinal data through recurrent neural networks, as well as solutions to explore a dataset through unsupervised learning approaches with autoencoders. These two architectures combine into a recurrent autoencoder which allows for the embedding of longitudinal data into a low dimensional, non-temporal representation while conserving the dataset's properties. Autoencoders can be as such interpreted as a form of non-linear principal component analysis on structured data. The subsequent embedding can then be used for a range of applications, such as visualization (by setting the embedding dimension to two or three, for instance), and cluster analysis through any well-known clustering procedure for multivariate, non-temporal data.

This paper investigates the use of such methods on a real dataset, and confronts its results to ones obtained with more traditional methods of longitudinal data clustering, namely the kml and proc traj procedures. Simulated datasets will also be used to point situations where deep learning could have a special interest.

Section 2 describes the autoencoder and recurrent neural networks architecture and precisely defines how they can lead to the clustering of longitudinal clustering. In section 3, the presented architecture will be used on both real-life and simulated datasets, along with the two other longitudinal clustering methods presented above. Results from each approach will then be compared to ensure consistency and robustness. Finally, section 4 concludes with a discussion on potential advantages and drawbacks of recurrent autoencoders for longitudinal clustering compared to more traditional approaches.

# Method

**Overall**

To investigate the potential application of the previously defined architecture in longitudinal data clustering, we decided to confront the results of two well established longitudinal data clustering procedure -the proc traj algorithm from SAS and the kml function from R-, against the proposed deep learning clustering architecture.

*kml and proc traj*

The kml procedure (1) is an adaptation of the K-mean clustering algorithm to longitudinal data. Traditional K-mean clustering is an iterative algorithm consisting of two phases. First, k points $(m_1,...,m_k)$ are randomly initialized in the data space. A k partition is then defined from these points by attributing each data point to the nearest ($m_i$), according to a predefined distance. The ($m_i$) are then updated to be their cluster mean. This step is then repeated until convergence of the cluster centers.

The kml procedure is essentially identical. The k points are replaced with randomly sampled sequences, and the distance is selected through the panel of existing distance for longitudinal data (eg. L1, L2, Fréchet distance, Dynamic Time Warping).

Proc traj (2) constitutes a different approach to the same problem, and consists of fitting a mixture of parametric models to the data through an Expectation-Maximization algorithm.

**Deep autoencoder for longitudinal clustering**

*Linear regression*

Linear regression is a widely used modelling technique consisting of modelling dataset with a linear, multivariate approximation of the real, studied phenomenon. This simple approach also shares similarities with regressive feedforward neural networks, and make as such a good introduction to understand them.

The objective of linear regression is to fit a set of N observations $\{(X_i, y_i)\}_{0<i<N+1}$ with a parametric, linear approximation in W such as:

$$\hat{y} = W_e \cdot X + b$$

$$\text{with } W_e = \underset{W}{\text{argmin}}(\frac{1}{N} \cdot \sum_{i=1}^{N}(\hat{y}_i - y_i))$$

The parametric solution $W_e$ can be determined either analytically, or through the use of an optimization algorithm such as gradient descent. Linear models are known to be easy to fit, as well as their good generalization behavior. They however fail to capture any non-linearity, as well as variable interactions.

*Feedforward neural networks*

Feedforward neural networks, are among the most widespread deep learning models (3). They can be used in a variety of modelling tasks, such as regression, where then can be seen as a nonlinear expansion to traditional mean square linear regression.

The idea behind feedforward neural networks is to fit a linear model to a transformed set of observations $\{(\Phi(X_i), y_i)\}_{0<i<N+1}$ where not only the model's parameters, but also $\Phi$, are learnt from the data.

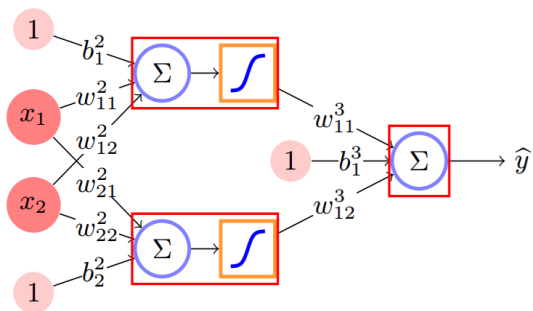

*Fig.1: Toy example of a feedforward neural network in regressive settings*

The traditional approach to enable a neural network model to learn non-linear interaction from the data is to inject linear combination of the investigated parameters into simple, nonlinear functions, whose outputs are then either used to perform a linear modelling task, or as inputs to be injected in another set of nonlinearities.

The toy network presented in figure 1 is an example of a feedforward neural network with one hidden layer, used in a regression setting. As in linear regression, the objective is to fit a set of N observations $((x_1, x_2)_i, y_i)_{0<i<N+1}$ with a parametric, non-linear approximation:

$$\hat{y} = w_{11}^3 \cdot f^1(w_{11}^2 \cdot x_1 + w_{12}^2 \cdot x_2 + b_1^2) + w_{12}^3 \cdot f^2(w_{21}^2 \cdot x_1 + w_{22}^2 \cdot x_2 + b_2^2) + b_1^3$$

While the parameters are typically obtained from the minimization of the square loss function, two differences arise from the introduction of parametric non linearities in the model:

- The parametric solution that minimizes the cost function cannot be expressed in closed form, and must be estimated through gradient based optimization
- The cost function's gradient is not straightforward to compute, but can be obtained using the backpropagation algorithm (4)
- The resulting optimization problem loses its convexity, and the warranty of finding the global minimum

The entire family of feedforward neural network can then be derived from this toy example, mainly through the variation of three hyperparameters:
- Number of composed function in chain (number of layer)
- Number of newly build feature per layer (number of neurons)

- Type of nonlinearity applied to the neurons output (i.e. choice of functions $f^1$ and $f^2$)

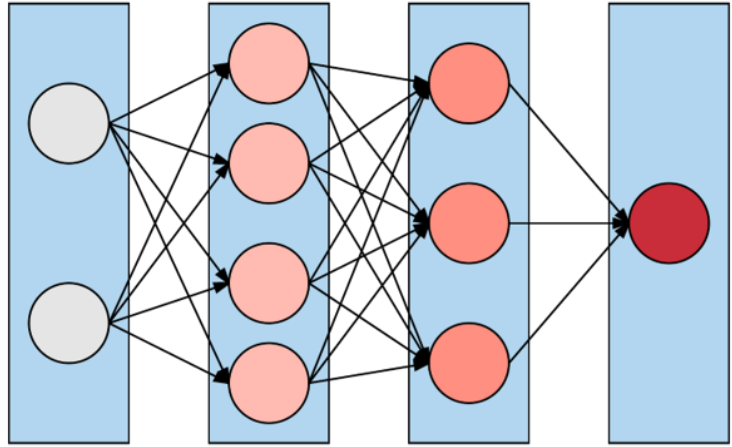

*Fig. 2. Neural network model with two hidden layers (in blue). The first hidden has six neurons, which are injected into the second layer's four neurons. The red and green layers are respectively the model's input (endogenous variables) and output (exogenous variables)*

Increasingly complex models can be built by tuning these hyperparameters, thus designing a model most fitted to the problem's complexity.

If theoretically, any type of nonlinearity $f$ can be applied to the neurons' outputs, practitioners usually choose between a limited number of options that have been empirically proven to perform well, with the two most notable examples being the hyperbolic tangent function and the linear rectified unit.

*Recurrent neural network*

Feedforward neural networks can in theory be used to analyze a sequence of data $x^{(1)}, x^{(2)},...,x^{(n)}$, for instance by feeding every timesteps directly in a neural network. However, this approach can turn out to be rather cumbersome. Indeed, trying to fit a densely connected feedforward network to a sequence of data results in an exponential increase in the number of parameter in the model with the increase of the data sequence's length. In addition, feedforward neural networks are clearly not able to model sequences of different lengths, which prevents their use in a number of longitudinal data analysis problem.

Recurrent neural networks are a family of neural network that specializes in the analysis of longitudinal data (3). The main idea behind the elaboration of a recurrent neural network is to devise a model that shares its parameter across all timestep within the data. Instead of feeding the whole sequence to a neural network, each timestep in the data is sequentially fed to the neural network, which is given a retroactive loop to enable him to account for past iterations.

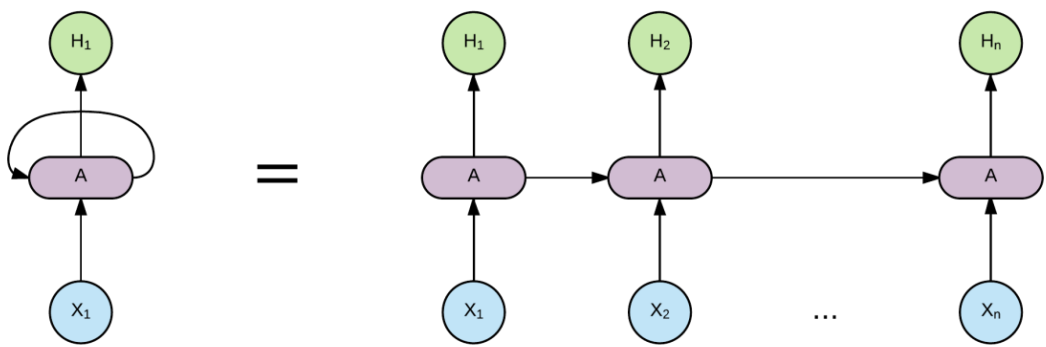

*Fig. 3. Recurrent neural network architecture, with on the right its looped form and on the left its unrolled representation*

This family of neural network can be used in a variety of settings, such as, for instance, non-linear regression on sequential data. The objective is to fit N observations $((X_1, ..., X_n)_i, y_i)_{1<i<N}$ with a parametric approximation such that:

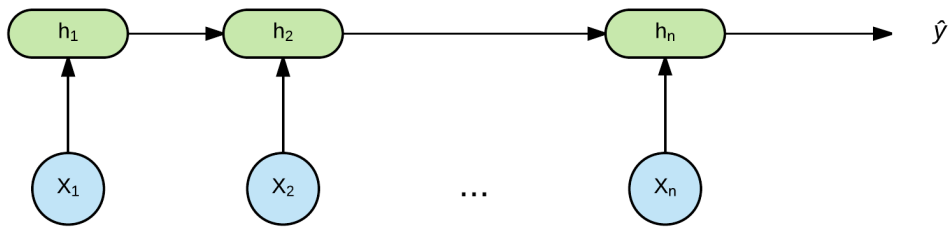

Fig. 4. Recurrent neural network for regression on sequential data

$\forall i \in \mathbb{N}, i \in [1, n]$

$h_i = f(x_i, h_{i-1})$    $with\ f\ a\ parametric\ family\ of\ functions$

$\hat{y} = W . h_n$

Similar to a feedforward neural network, $\hat{y}$ is finally approximated from the recurrent neural networks last output layer through the fitting of a traditional linear regression, whose parameter are obtained by the optimization of a least square cost function on both W and the parameter of the parametric function f, which was historically chosen to be a feedforward neural network.

However, the use of a feedforward neural networks for the f function leads to several issues in the model fitting process. Although these issues outreach the scope of this article, the interested reader will find their extensive description in (5).

Several architectures have been devised to counter these problems, such as the Gated Recurrent Unit, or The Long-Short-Term-Memory (LSTM) cell, the latter being one of the most wildly used, and the architecture chosen for experimentation in this paper.

The Long-Short-Term-Memory cell (6) is a variation of a traditional recurrent network, and defines the parametric function f defined for the proposed approach to regression on sequential data as:

$$f_t = \sigma_g(W_f \cdot x_t + U_f \cdot h_{t-1} + b_f)$$

$$i_t = \sigma_g(W_i \cdot x_t + U_i \cdot h_{t-1} + b_i)$$

$$o_t = \sigma_g(W_o \cdot x_t + U_o \cdot h_{t-1} + b_o)$$

$$c_t = f_t \circ c_{t-1} + i_t \circ \sigma_g(W_f \cdot x_t + U_f \cdot h_{t-1} + b_f)$$

$$h_t = o_t \circ \sigma_h(c_t)$$

With:

- $x_t$ as the input vector
- $h_t$ as the output vector
- $c_t$ as the cell state vector
- W, U, and b the cell's parameter matrices and vector
- $\sigma_g$ the logistic function
- $\sigma_c$ and $\sigma_h$ the hyperbolic tangent function

Although this set of equation defined the original LSTM cell as it was first expressed, several variations on this architecture have since been devised, such as the peephole LSTM.

*Undercomplete Autoencoder*

Both recurrent and feedforward neural network are defined as supervised learning algorithms. The model is built by setting its parameters in order to optimize its ability to approximate the model's exogenous variables. However, clustering algorithms are typical examples of an unsupervised setting, in which a structure within the observed data is to be extracted.

Autoencoders are a family of neural networks that adapts the concepts of supervised learning through feedforward and recurrent networks to enable their use in unsupervised settings.

The objective of autoencoders is to use neural networks' ability to define complex, nonlinear interaction to embed a dataset into a richer representation, which can then be used to perform several tasks such as dimensionality reduction, data compression or visualization.

The study of autoencoders is a growing area of research, that lead to the elaboration of entire families of algorithms, such as the undercomplete autoencoder.

An undercomplete autoencoder is built from two separate neural networks, an encoder *f* and a decoder *g*. The encoder takes a set of N observations $\{X_i\}_{0<i<N+1}$ as inputs, and outputs for each of them a vector $h_i$ of dimension smaller than its entry.

The decoder then takes the $h_i$s as entries and tries to rebuild the $X_i$s. The two parts of the autoencoder are jointly optimized in order to minimize a reconstruction error:

$$L(X, g(f(X)))$$

L is typically chosen as the mean squared error (3).

To perform the reconstruction task, the encoder, forced to compress the data in a smaller representation, will need to capture the data's most informative characteristics (3).

As such, an undercomplete autoencoder can be seen as an expansion of kernelized principal component analysis in which the kernel function is learnt from the dataset. In fact, an undercomplete autoencoder with linear decoder and mean squared error cost function will learn the same data decomposition as a linear principal component analysis (3).

*Recurrent Autoencoder and longitudinal clustering*

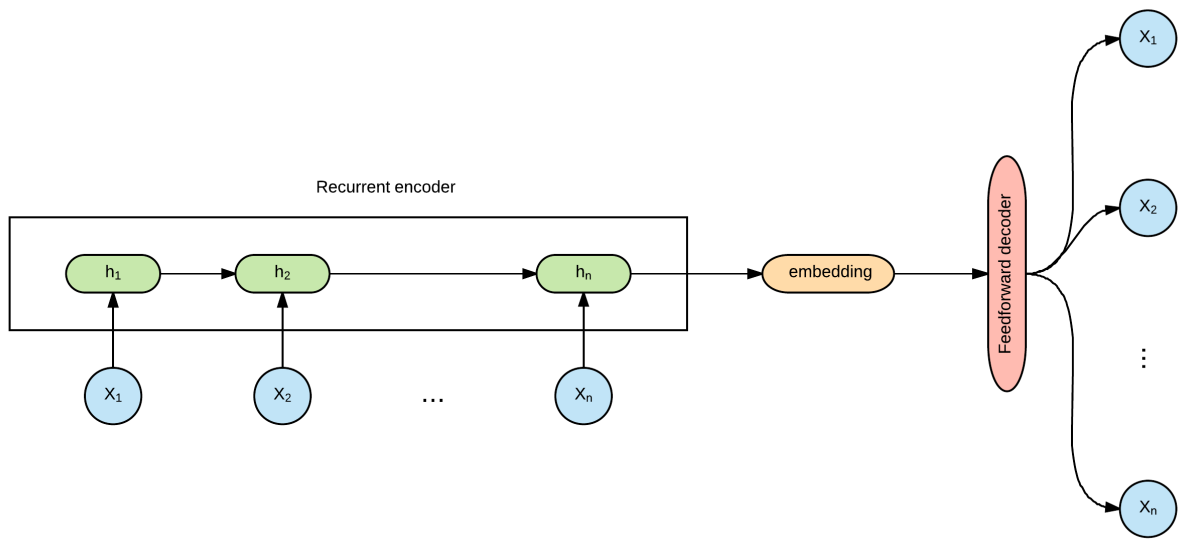

*Fig. 5. Neural network used to perform longitudinal clustering: A recurrent neural network "reads" the studied longitudinal data and outputs a low dimensional vector.*

*This vector is used as an entry for a multilayer perceptron, which then reconstructs the original longitudinal data*

While undercomplete autoencoders are traditionally used with feedforward neural networks, they can be expanded to the analysis of longitudinal data, for instance by using a recurrent neural network as the encoder (7). The extracted embedding can then be used as a low-dimensional, non-temporal representation of the studied sequential data, on which traditional, multivariate clustering algorithms can be applied. Several architectures are available to the practitioner, among them the one described in fig. 5, selected to perform the experimentation reported in this paper:

- A recurrent neural network in the setting described above is used to sequentially read the data sequences
- The last hidden layer is converted into a low dimensional vector with a linear, feedforward neural network
- The low dimensional representation is fed to a feedforward neural network
- The feedforward neural network's output is then used to copy the data sequence through multiple linear regression (as much as there is values in the data sequences)
- All the model parameters, from the recurrent network, the feedforward networks and the linear regressions, are tuned by the optimization of a least square cost function
- The cost function is optimized with a variant of the stochastic gradient descent with momentum algorithm (RMSProp (8))

- The function's gradient is computed with the backpropagation algorithm

- The low dimensional vector obtained from the recurrent network's last hidden layer is used to perform visualization and multivariate data clustering.

**Data analysis**

*Proc traj*

The proc traj based data partition was retrieved from the results of the original study (9) . In this reference, the number of clusters was fixed to six, value obtained through the optimization of a parsimonious index. It is noticeable that, in the present paper, the number of clusters retained for all three approaches was also fixed to 6 to ensure good comparability.

*kml*

The partition obtained with longitudinal distance based K-mean was obtained using the related R package kml. The algorithm was ran 20 times with random initialization and the solution with maximum Calinsky-Harabasz criterion was selected for confrontation with the other procedures.

*Recurrent Autoencoder*

The partition obtained with longitudinal distance based K-mean was obtained using the related R package kml. The algorithm was run 20 times with random initialization and the solution with maximum Calinsky-Harabasz criterion was selected for confrontation with the other procedures. The recurrent autoencoder was manually implemented in Tensorflow (10), a dedicated python based library for deep learning. The selected architecture was defined as afordescribed. Hyperparameters (architecture of the recurrent encoder, number of neurons in the encoder, embedding dimensionality, number of layers in the decoder, number of neurons per layers in the decoder) need to be fine-tuned by a deep-learning practitioner. This complex process, on which the interested reader will be able to find material in reference (3), still constitutes an active area of research, for which no wildly agreed upon methodology is known, and as such far exceeds the scope of this article. For purpose of visualization, the embedding's dimensionality was set to two.

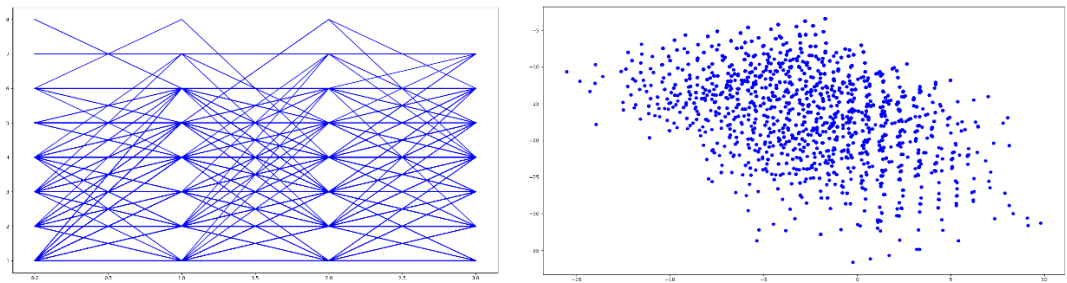

*Fig. 6. Left: Longitudinal plot of the dataset's trajectories, Right: Plot of the corresponding bi-dimensional embedding*

Once a satisfying bi-dimensional, atemporal embedding is extracted from the dataset, clustering was then extracted from it by traditional, Euclidian distance based, K-mean clustering. Similarly, as for kml clustering, 20 partitions where extracted from random initialization, and was kept the best cluster set in regard to the Calinsky-Harabasz criterion.

*Coherence of the three partitions*

To establish quantifiable coherence between the three partitions, posterior probabilities of an individual to belong to a given cluster were computed for the three clustering procedures:

- As a gaussian mixture based model, the proc traj procedure gives us natural access to the desired probabilities
- The kml procedure offers a built-in cluster membership probability tool
- The auto-encoding based clustering membership probabilities were computed on the extracted embedding, under an assumption of gaussian prior for each cluster, with mean and covariance computed on the cluster's members

The extracted probabilities' correlation matrix is finally represented graphically using a spherical representation (11).

## Results

**Real-data application**

*Dataset*

The three aforementioned clustering procedures were confronted against each other on a dataset obtained from the E3N cohort study (9), which describes the evolution of 80110 women's body shape across time. Measurements consists of an 8 levels ordinal scale representing "extreme thinness" to "obesity", and were established for 4 times of the subject's life:

- Around the age of 8 years
- At menarche
- At ages 20 to 25 years
- At ages 35 to 40 year

Longitudinal data clustering of these scale measurements has been used in a series of papers with proc traj. It is of interest to ensure this clustering is retrievable with the kml algorithm and the autoencoder as well.

*Cluster confrontation*

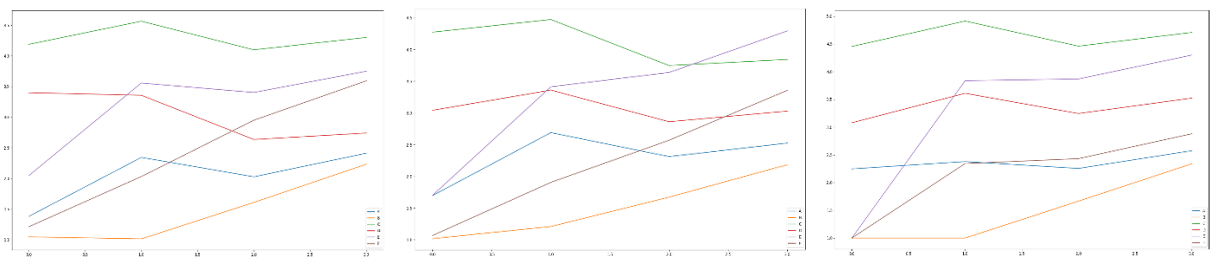

*Fig. 7. Cluster Mean trajectories for each procedure (from left to right, autoencoder based clustering, kml, proc traj)*

As shown in fig. 7, clusters mean trajectories obtained are similar enough to enable a potential cluster association within each clustering method. Although encouraging, this result alone only constitutes an empirical observation and cannot assert any satisfying conclusion on its own.
To establish quantifiable coherence between the posterior probabilities of an individual to belong to a given cluster within a given method, a spherical representation of the correlation matrix of these posterior probabilities is displayed in fig.8. It indicates that cluster membership probabilities among the three approaches are indeed correlated, thus ensuring stability within the investigated procedures.

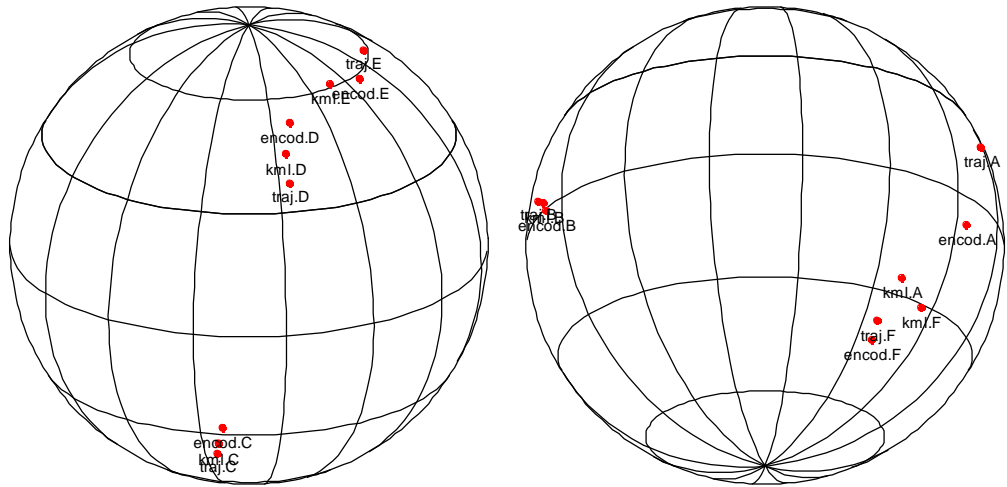

*Fig. 8. Spherical projection of cluster belonging probabilities' correlation matrices. Traj, encod and kml points respectively link to clusters obtained with the proc traj, autoencoder and kml procedures. Letters link to visually similar clusters in the different approaches. For X = A, B, C, D or E encod.X, kml.X and traj.X are very close, this indicates that the clustering given by the 3 studied approaches are similar.*

**Simulated data application**

*Dataset*

To illustrate situations where the autoencoder outperforms traditional clustering methods, the kml and autoencoder procedures where confronted against each other on a simulated dataset corresponding to a hypothetical study of quality of life measurements amongst cancer patients. Patients are divided into 2 groups corresponding to 2 treatment allocations:

- Group A patients receive a treatment every 4 month. Their quality of life falls once treatment is administered because of side effects, gets up because of treatments effects, then stabilize because treatment is no more efficient, then comes another sequence of treatment thus leading to a sine shaped evolution of the patient's quality of life
- Group B patients receive continuous treatment, leading to stable life quality

The quality of life evaluation at time *t* was modeled such that:

$$QoL_A(t) = 5 \cdot \sin\left(\frac{\pi}{2} \cdot t + U(-2, 2)\right) + 10 + N(0, 1)$$

$$QoL_B(t) = 10 + N(0, 1)$$

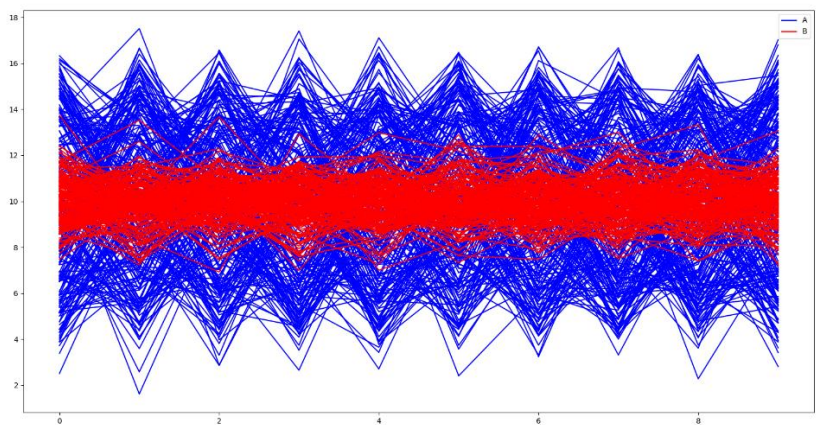

*Fig. 9. Longitudinal plot of the simulated dataset. Patients in group A (in blue) have a sine shape evolution while patients from group B (in red) are stable.*

*Cluster confrontation*

The methodology used to derive both clustering is as aforedescribed. For the kml procedure, the number of clusters was chosen to maximize the Calinski-Harabasz index. Proc traj was not used here because it is not designed to tackle periodic trajectories and so is not adapted to this type of data. As can be seen in fig. 8, the kml procedure fails to identify the two different groups, and prefers it a partition of 7 clusters. However, it does manage to correctly group together individuals from the group B, and as such is still able to bring some insight about the dataset to the practitioner.

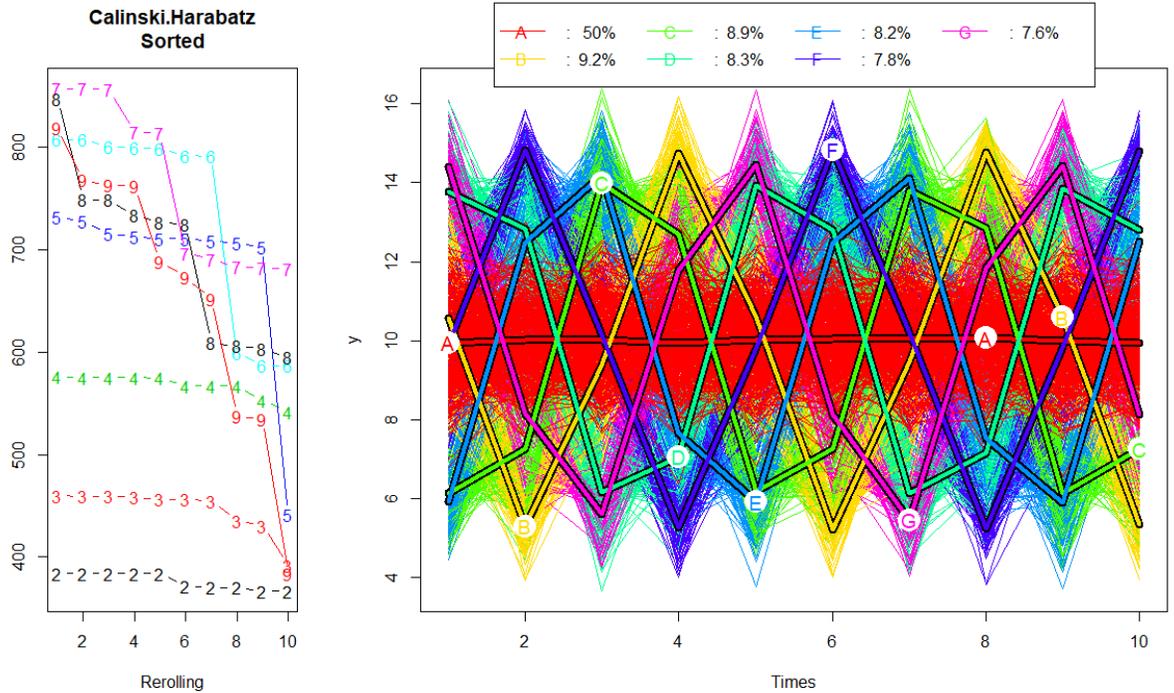

*Fig. 10. Left : display of the Calinski-Harabasz indices for all ran clustering procedures, Right: Best extracted 7-partition, with mean cluster curve in bold and each individual subject grouped according to cluster belonging by color*

Fig. 9 shows a plot of the autoencoder clustering procedure's extracted embedding. Groups A and B are clearly separated into two easily identifiable, non-spherical clusters. Said clusters can then be extracted through the use of different methods, such as spectral, or agglomerative clustering algorithms.

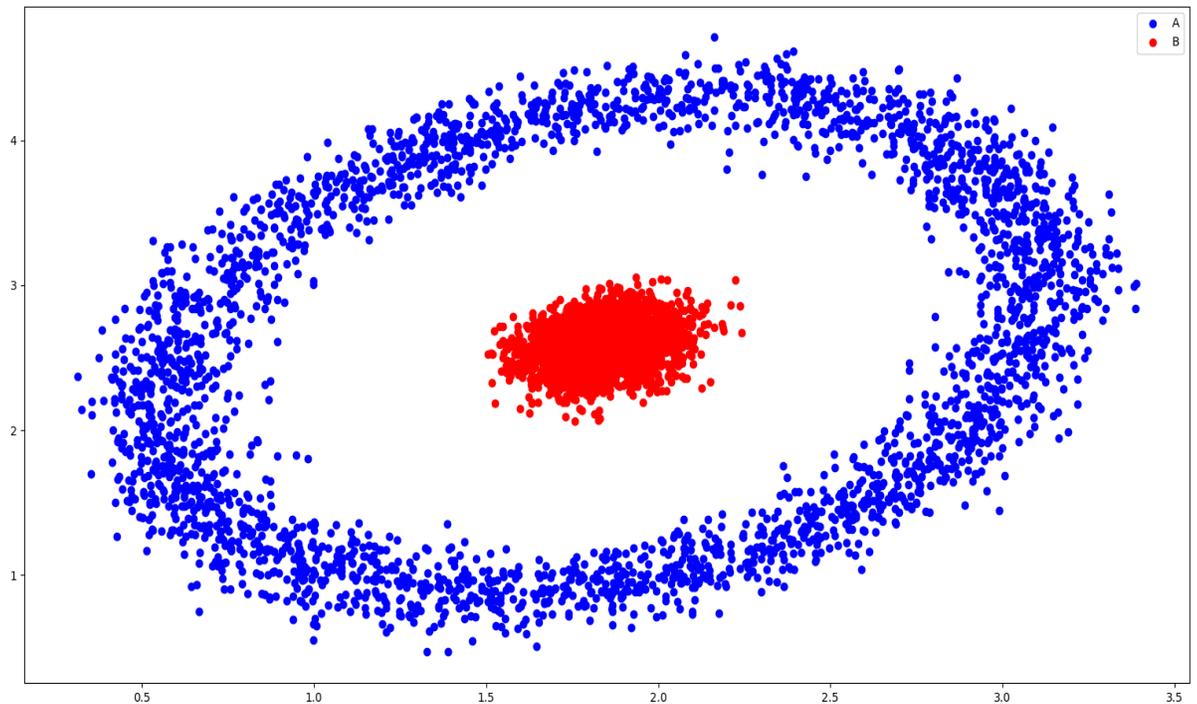

*Fig. 11. Bi-dimensional, atemporal embedding of the simulated dataset, with in blue embeddings corresponding to group A patients' curves, and in red embedding corresponding to group B.*

Although the kml procedure managed to identify interesting patterns in this experiment, it was clearly outperformed by the autoencoder based approach. Indeed, the two subject groups crafted for the simulation are only correctly extracted from the bi-dimensional embedding provided by the recurrent autoencoder. One might object that K-means based procedure are not suited to the analysis of non-spherical clusters which the simulated dataset is clearly made of, and that other longitudinal clustering approaches might be better suited to the studied problem. However, with this experiment and the aforementioned one on the E3N dataset, the proposed autoencoder based approach was shown to perform correctly in both spherical and non-spherical clusters setups, thus proving its adaptability. It is also worth mentioning that the embedding itself, in this experiment, allows us to visualize useful information on the studied dataset. Indeed, the extracted non-spherical embedding might illustrate non-spherical relationships happening in the corresponding longitudinal dataset. Although not theoretically backed, this kind of insight can turn out to be useful, for instance by guiding the practitioner away from the use of spherical cluster based statistical methods such as K-Means or Gaussian mixture based procedure, as well as cluster mean curve plotting as graphical results displays.

## Conclusion

Cluster analysis of longitudinal data is an important topic for epidemiologists because it provides insight in data not achievable with classical methods among which random effect generalized linear models for repeated measures. Several approaches exist at the moment, among which mixture modelling and historical segmentation algorithms like K-Means. This paper shows that deep learning technics could be a promising alternative, being able to grasp non-spherical clusters (where K-Means procedure is not efficient) with complex temporal evolution or non-Gaussian mixtures (where standard mixture modelling shows strong limitations).

Deep neural networks can suffer from several drawbacks that render them impractical in a number of cases. Indeed, these methods are known to perform poorly on small datasets. In addition, they require careful implementation, and hyperparameter tuning, which often makes them difficult to implement to non-specialists. However, these methods constitute powerful tools that have already achieved state of the art results on complex modelling tasks, such as image, voice or text modelling and could be thus promising for epidemiological data.

Coherence between the three approaches is ensured for simple cases, this is likely to comfort results already published even if it is strongly suggested to researchers to look for robust segmentations of trajectories with a systematic use of different clustering approaches.

The proposed network architecture leaves several questions that might lead to future works. First, the investigated method separates the embedding extraction and the clustering procedure into two, distinct steps lead by two distinct optimizing processes. Although leading to satisfying results, such an implementation stands out of the usual deep learning practitioner's philosophy. Indeed, it is usually preferred, when designing a neural network for a given task, to gather every part of the procedure in a single optimization process. Indeed, the low dimensional embedding extracted from the data in the proposed approach isn't necessarily optimized for a clustering task to be performed subsequently. The design of architectures optimizing jointly the low dimensional embedding and the extraction of a data partition is an active area of research, with some promising methods already proposed(12).

At last, this paper clearly shows that technics originally coming from computer science have a great potential in the field of epidemiology and biostatistics. These new approaches are most often quite demanding in terms of sample size requirements, but they are the most capable to grasp non-linear and complex relationships, with potential breakthroughs unthinkable until now.

# Informations

Louis Falissard designed the neural network architecture, proceeded to their implementation and their deployment on the two studied datasets, and wrote the presented article.

Guy Fagherazzi provided the E3N dataset along with the clustering results extracted from the proc traj procedure obtained for a previous study.

Newton Howard acted as academic supervisor for student Louis Falissard

Bruno Falissard developed the studies' methodology, and defined the simulation dataset

E3N data can be made available upon request to the corresponding author